# QuesGenie: Intelligent Multimodal Question Generation

Ahmed Mubarak
*Faculty of Computer and Information Sciences*
*Ain Shams University*
Cairo, Egypt
ahmedmubarak.hsin@gmail.com

Amna Ahmed
*Faculty of Computer and Information Sciences*
*Ain Shams University*
Cairo, Egypt
amna.a.mirghani@gmail.com

Amira Nasser
*Faculty of Computer and Information Sciences*
*Ain Shams University*
Cairo, Egypt
amira20nasser@gmail.com

Aya Mohamed
*Faculty of Computer and Information Sciences*
*Ain Shams University*
Cairo, Egypt
ayahmohamed1839@gmail.com

Fares El-Sadek
*Faculty of Computer and Information Sciences*
*Ain Shams University*
Cairo, Egypt
fares7elsadek@gmail.com

Mohammed Ahmed
*Faculty of Computer and Information Sciences*
*Ain Shams University*
Cairo, Egypt
mohammed24266@gmail.com

Dr.Ahmed Salah
*Faculty of Computer and Information Sciences*
*Ain Shams University*
Cairo, Egypt
ahmed_salah@cis.asu.edu.eg

TA Youssef Sobhy
*Faculty of Computer and Information Sciences*
*Ain Shams University*
Cairo, Egypt
youssef.sobhy@cis.asu.edu.eg

*Abstract*—In today's information-rich era, learners have access to abundant educational resources, but the lack of practice materials tailored to these resources presents a significant challenge. This project addresses that gap by developing a multimodal question generation system that can automatically generate diverse question types from various content formats. The system features four major components: multimodal input handling, question generation, reinforcement learning from human feedback (RLHF), and an end-to-end interactive interface. This project lays the foundation for automated, scalable, and intelligent question generation, carefully balancing resource efficiency, robust functionality and a smooth user experience.

*Keywords—automatic question generation, multimodal question generation, reinforcement learning, education*

## I. Introduction

Creating assessment questions is a time-consuming and labor-intensive task for educators. Traditional methods require manual extraction of information from materials, which can lead to inconsistencies and errors. Additionally, students often struggle to find varied practice questions that cover all aspects of the material they are studying. With the increasing use of multimedia in educational content, there is a growing need for systems that can process various data types, including text, diagrams, and audio recordings. This project aims to address these challenges by developing a system that automates the generation from multiple input sources, thus reducing the workload on educators while providing students with a valuable tool for exam preparation.

The project aims to achieve four main objectives. The first is to process various input formats (namely presentation slides, PDF documents and audio recordings) to serve as sources for question generation. The second is to generate different types of questions (true/false, multiple choice questions or MCQ, fill-in-the-blanks and matching) from both the text and images present in the input sources. Thirdly, human feedback, along with LLM evaluation, will be used to improve the model through reinforcement learning. Finally, we aim to achieve all of this through a robust end-to-end system, allowing users to access it through a seamless and intuitive mobile application.

## II. Related Work

### A. Question Generation

Automatic Question Generation (AQG) [1] is tasked with generating questions from given input, which is usually either text or an image, and generating an answer for the question. There are a few classifications that can be used to categorize AQG problems. The questions generated can either be closed-domain, which are limited in scope to a specific domain of knowledge, or open-domain, which are general and not restricted to a particular domain. The questions themselves can be of different types, including: factual questions, which are simple, objective and typically have one-word answers; multiple-sentence questions, whose answers span multiple sentences; boolean questions with yes/no answers; and deep understanding questions, which require inference from multiple facts. Several approaches exist for AQG, ranging from approaches that rely purely on natural language processing (NLP) to state-of-the-art Transformers [2] and Large Language Models (LLMs). AQG is evaluated using different metrics, which can be automatic or human-based. Examples of automatic metrics are BLUE and ROUGE, commonly used across NLP tasks. Examples of human-based metrics are relevance, difficulty, answerability, fluency, and so on.

Visual Question Generation (VQG) is a subset of AQG that focuses on question generation from images. This problem aims to generate the questions from the visual elements of the image, but can also supplement it with text that is either captured from the image directly or provided as a caption or description. This project tackles AQG and VQG. Text-based questions are open-domain, while image-based questions are limited to charts and diagrams.

The generated questions are either factual or boolean questions. To evaluate the generated question-answer pairs, we use BLEU [3] and ROUGE [4] scores.

*B. Reinforcement Learning*

Reinforcement Learning (RL) is the process by which a model improves by taking action to maximize a reward signal. A variant of RL is Reinforcement Learning from Human Feedback (RLHF), which aligns Language Models with human preferences by using a reward model that approximates human feedback. It involves three broad steps: developing the Language Model, developing a reward model based on human feedback, and finally, optimizing the language model using RL algorithms such as Proximal Policy Optimization (PPO) [5]. However, human feedback can be difficult to obtain in some cases, in which case Reinforcement Learning from AI Feedback (RLAIF) can be used. This substitutes human feedback for feedback generated using other Language Models.

Proximal Policy Optimization (PPO) is a reinforcement learning algorithm that has proven effective and stable for fine-tuning language models. PPO balances between improving the model's performance and avoiding large, destabilizing updates. In RL, an agent learns to take actions that maximize a reward. PPO improves this process by using a mathematical technique that updates the model's policy (the way the model generates output) without making large changes in a single step, leading to smoother and more stable updates during training.

### III. METHODOLOGY

Fig. 1 illustrates the system architecture. The input sources for question generation are parsed according to their format (either PowerPoint, PDF or audio) and converted into chunks, which are either text-based or image-based. The image chunks are classified as diagrams or not, and only the diagrams are used for visual question generation. The VQG model generates a question-answer pair from each provided diagram. The text chunks are preprocessed and then passed to the next module, where key terms are extracted using TF/IDF. Both the text segments and extracted key terms are embedded. We then calculate the similarity between the key terms and each segment to retrieve the top-k most relevant segments for each keyword, which will be used as sources for question generation

Question and answer pairs are generated in different formats: MCQ, true/false, fill in the blanks, and matching questions. The user then optionally provides feedback, which is collected and stored in the system. This feedback will later be used to enhance and optimize the model through reinforcement learning. Each of the modules mentioned will be discussed in detail in the following sections.

*A. Input Preprocessing*

Before generating any type of question, the system begins by extracting and cleaning the input context. This context may come from documents, slides, or audio files, and it forms the foundation for all subsequent question generation phases. The audio is transcribed using Whisper [6] and converted to Audio Chunks that contain the text transcript along with start and end timestamps. Both PowerPoint and PDF documents are processed to result in two types of output: Image Source, containing the image file path and its location in the source, and Text Chunk, containing the text and its location in the source (the location is the page number for PDF and slide number for PowerPoint). At the end of this step, the input has been converted into chunk objects, containing the text that will serve as the question source, as well as the metadata linking the question back to its location in the source file. Fig. 2 shows the outputs of the preprocessing step for various input types. These chunks are then passed to the next phases.

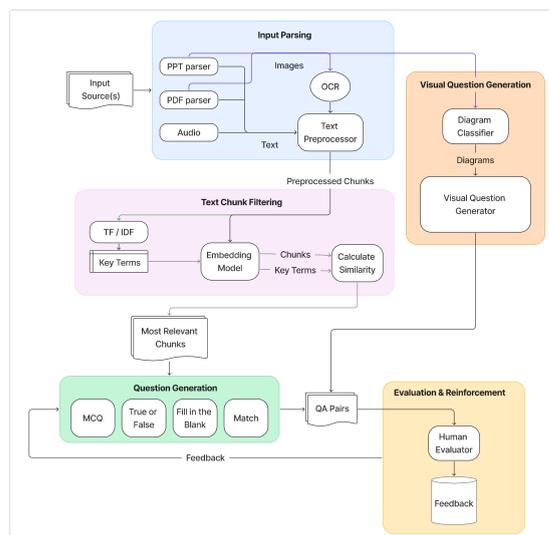

**Fig. 1.** Schematic representation of the project architecture. Input sources are parsed to yield text and images. Images are classified and diagrams are sent for VQG . Text is preprocessed, and keywords are extracted and used to retrieve the most relevant chunks. These are then used to generate various types of question-answer pairs, which are rated by users. This feedback is later used to reinforce the model.

*B. Text-Based Question Generation*

We chose T5 (Text-to-Text Transfer Transformer) [7] as our question generation model for both MCQ and true/false question types. For generating question-answer pairs, we used the SQuAD (Stanford Question Answering Dataset) [8], which contains context paragraphs from Wikipedia along with corresponding questions and answers. To generate plausible distractor options for MCQs, we used the RACE (ReAding Comprehension from Examinations) dataset [9], consisting of English exam questions for school students, each including a question, a correct answer, and several distractors.

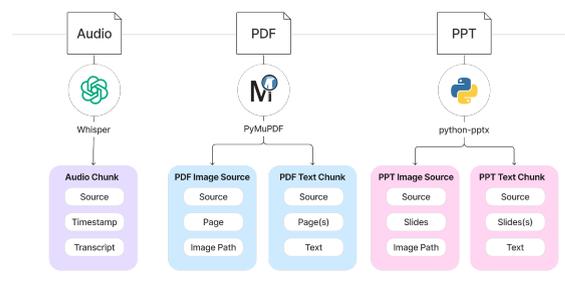

**Fig. 2.** Input preprocessing for the various input types. Every type is mapped to a chunk object containing the source for question generation (text context or image), along with metadata for tracing them back to their location in the source

This dataset is particularly useful for training models to generate semantically similar but incorrect alternatives. Finally, for generating true/false questions, we utilized the BoolQ (Boolean Questions) dataset [10]. BoolQ contains natural yes/no questions paired with short paragraphs from Wikipedia. Table I provides details about the datasets.

TABLE I. DATASETS FOR TEXT-BASED QUESTION GENERATION.

| Dataset | Train Size | Validation Size | Test Size |
|---------|-----------|-----------------|-----------|
| SQuAD | 75,719 | 10,570 | 11,877 |
| RACE | 87,866 | 4,887 | 4,934 |
| BoolQ | 9,427 | – | 3,270 |

The last two question types are relatively straightforward to generate. Matching questions are created by generating a few question-answer pairs and then shuffling the answers. Fill-in-the-blank questions are generated using a different approach, using the Python Keyphrase Extraction (PKE) toolkit to identify the most important words or phrases from the context [11]. These key terms are then replaced with blanks, forming fill-in-the-blank questions

### C. Visual Question Generation

The visual question generation pipeline consists of two main steps: diagram classification and question generation.

To effectively tackle the task diagram classification, we constructed a specialized dataset comprising two classes: 'Diagram' and 'None'. We ensured that the 'none' class included drawings and images with text that were not diagrams, to avoid training a model that automatically associated any text or illustration-type visuals with diagrams. Fig. 3 demonstrates samples from the datasets used, and Table II summarizes the number of samples used from each.

TABLE II. SUMMARY OF THE DATASETS FOR DIAGRAM CLASSIFICATION

| Class | Training Dataset | Testing Dataset |
|-------|------------------|-----------------|
| Diagram | 7,000 samples from ChartQA and 3,000 samples from JasmineQiuqiu/ diagrams_with_captions_2 | 3,000 samples from ViRFT_COCO_base65, 2,000 samples from TextOCR-Dataset and 5,000 samples from pl-text-images-new-5000 |
| None | 3,000 samples from linbojunzi/diagram_images | 3,000 samples from HuggingFaceM4/COCO |

We used this dataset to fine-tune ResNet-50, optimizing for using balanced cross-entropy loss. We altered the weights of the classes, giving a weight of 1 to the diagram class and 3 to the none class, with the aim of increasing the model's precision. We call our model DiaClass.

Then, we pass the images classified as diagrams to the visual question generation model to generate question-answer pairs. We choose ChartGemma [12] as our model, which consists of a PaliGemma [13] backbone fine-tuned on datasets including synthetic charts, curated real-world charts, Wikipedia charts and in-the-wild charts. We used the following prompts to generate visual questions.

- **Description generation:** "<image> Generate a full description about this diagram."
- **Question generation:** "<image> Generate a question on this chart."
- **Question answering:** We input the question for the model to answer.

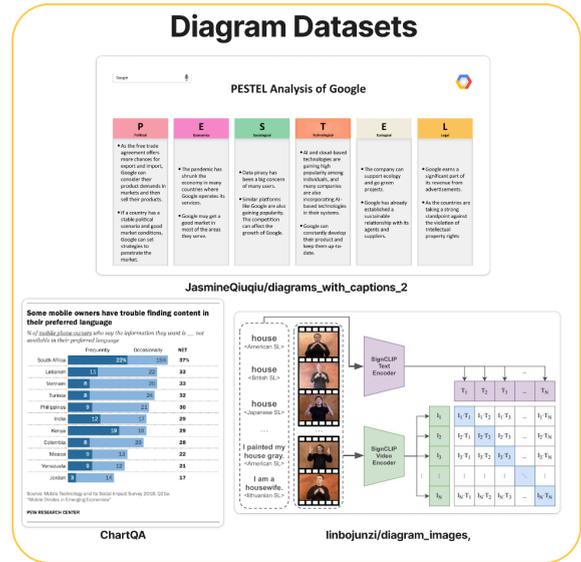

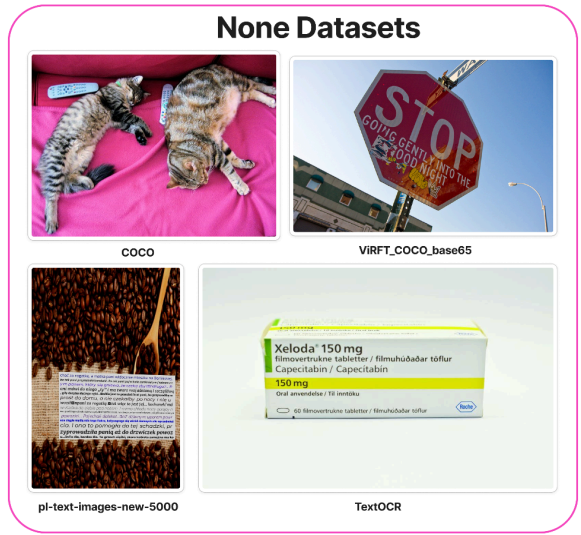

**Fig. 3.** Examples from the datasets used for diagram classification, visually separated as "diagram" or "none".

### D. Reinforcement Learning

The main objective of this phase is to use feedback scores to align the question generation with human preferences. We have limited the scope of reinforcement learning in our project to the T5 model used to generate text-based question-answer pairs. The system is designed to collect user feedback in the form of a 5-star rating measuring the generated question's relevance to the context. A single feedback sample is stored as an integer from 1 to 5, 5 being the best possible rating. The feedback dataset consists of the context, question, answer and rating, and will be used to train the reward model. We discuss the reward model in detail in the following section. To update the model, we use Proximal Policy Optimization (PPO) [5], which is a stable reinforcement learning algorithm suitable for sequence generation tasks.

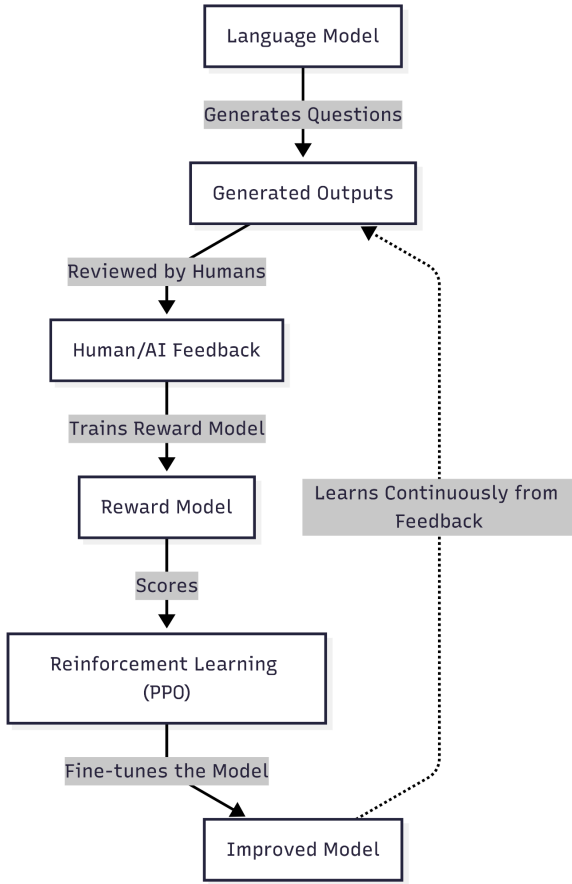

Fig. 4. Reinforcement learning process for AQG. First, the model generates questions, which are then reviewed by either human users or AI agents. This feedback is used to train a reward model to output scores that are used to reinforce the model through PPO.

## IV. RESULTS

### A. Text-Based Question Generation

We have fine-tuned two instances of T5 [7] to generate MCQs. The first model is T5-base fine-tuned on SQuAD, and is used to generate question-answer pairs. Table III summarizes the results for question-answer pair generation using T5. The model achieved strong performance on answer generation, indicating high overlap and semantic similarity with reference answers. For question generation, the model's results are within a reasonable range given the linguistic diversity of natural questions. These results demonstrate the model's effectiveness in generating coherent and contextually relevant question-answer pairs. Table IV compares some sample generated question-answer pairs with the reference pairs in the dataset for the same context.

TABLE III. RESULTS OF T5-BASE FOR GENERATING QUESTION-ANSWER PAIRS

| Metric | Generated Questions | Generated Answers |
|---|---|---|
| BLEU-4 | 0.1230 | 0.3150 |
| ROUGE-L | 0.3957 | 0.7426 |

TABLE IV. SAMPLE OUTPUT QUESTION-ANSWER PAIRS

| Context | Generated | Reference |
|---|---|---|
| There are a vast range of commodity forms available to transform a pet dog into an ideal companion. […] While dog training as an organized activity can be traced back to the 18th century, in the last decades of the 20th century it became a high profile issue. […] | **Question**: When was dog training first practiced?<br><br>**Answer**: 18th century | **Question**: In what type of plant cells does photosynthesis occur?<br><br>**Answer**: 18th century |
| The city and surrounding area suffered the bulk of the economic damage and largest loss of human life in the aftermath of the September 11 […] The World Trade Center PATH station, which opened on July 19, 1909 as the Hudson Terminal, was also destroyed in the attack. A temporary station was built and opened on November 23, 2003. […] | **Question**: When did the World Trade Center PATH station open?<br><br>**Answer**: July 19, 1909 | **Question**: On what date did the World Trade Center PATH begin operation?<br><br>**Answer**: July 19, 1909 |

The second model we used for MCQ generation is T5-large fine-tuned on RACE, and is used to generate the distractors (plausible but incorrect choices) that make up the multiple choices of the MCQ. We initially fine-tuned the base model for distractor generation but achieved unsatisfactory results, so we switched to the large variant. This resulted in much higher distractor quality. Table V summarizes our results, and a few sample outputs are shown in Table VI.

TABLE V. RESULTS OF T5-LARGE FOR GENERATING DISTRACTORS

| Metric | Distractor #1 | Distractor #2 | Distractor #3 |
|---|---|---|---|
| BLEU-4 | 0.1753 | 0.1177 | 0.0941 |
| ROUGE-L | 0.2653 | 0.2534 | 0.2597 |

TABLE VI. SAMPLE GENERATED DISTRACTORS

| Questions-Answer Pair | Generated Distractors | Reference Distractors |
|---|---|---|
| **Question**: According to the passage, if a family of origin has passion for literature, the members of the family will probably _____.<br><br>**Answer**: enter the field of literature | 'enter the stock market'<br><br>'be hesitant to put their money behind' | 'teach literature'<br><br>'hate literature'<br><br>'write poems' |
| **Question**: Using context clues we may infer that iodine, fluoride and calcium are _____.<br><br>**Answer**: substances which act positively or negatively on man's health | 'substances which provide breeding places for pests like mosquitoes or rats'<br><br>'substances which are found in the water supply' | "harmful substances in the water supply'<br><br>'substances which help provide breeding places for pests'<br><br>'substances contributory to good health' |

For generating true/false questions, we fine-tuned T5-base on BoolQ. Table VII summarizes the results of generating true/false questions using T5., and Table VIII compares sample questions generated by the model to the questions in the dataset generated from the same context.

TABLE VII. RESULTS OF T5-BASE FOR GENERATING TRUE/FALSE QUESTIONS

| Metric | Generated Questions |
|---|---|
| BLEU-4 | 0.2431 |
| ROUGE-L | 0.5647 |

TABLE VIII. SAMPLE GENERATED TRUE/FALSE QUESTIONS

| Context | Generated | Reference |
|---|---|---|
| Street Addressing will have the same street address of the post office, plus a ``unit number'' that matches the P.O. Box number. […] Therefore, for P.O. Box 9975 (fictitious), the Street Addressing would be: 1598 Main Street Unit 9975, El Centro, CA. Nationally, the first five digits of the zip code may or may not be the same as the P.O. Box address, and the last four digits (Zip + 4) are virtually always different. […] | Is a zip code the same as a post office? | does p o box come before street address |
| The new locks opened for commercial traffic on 26 June 2016, […]. The original locks, now over 100 years old, allow engineers greater access for maintenance, and are projected to continue operating indefinitely. | Are the original panama canal locks still in use? | Are the original panama canal locks still in use? |

Fig. 5 shows a sample fill-in the blank question generated by identifying keywords and replacing one of them with a blank.

### B. Visual Question Generation

Table IX summarizes the results of our diagram classification model. It achieves near-perfect accuracy with techniques like blurring the text in images, and adding non-diagram images that contain text in them for more robust classification

TABLE IX. RESULTS OF DIACLASS FOR DIAGRAM CLASSIFICATION

| Accuracy | Precision | Recall |
|---|---|---|
| 0.99 | 0.99 | 0.99 |

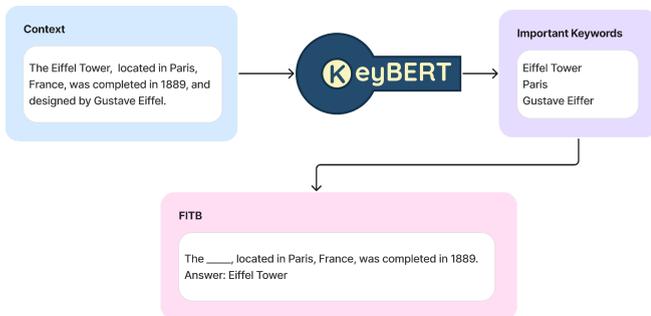

Fig. 5. Generating FITB questions by extracting keywords from the context and replacing them with a blank.

Table X shows a few sample question-answer pairs generated by ChartGemma.

TABLE X. SAMPLE VISUAL QUESTION-ANSWER PAIRS

| Image | Generated Question-Answer Pair |
|---|---|
| Fig. 6. A diagram of the DNS address resolution process. | **Question**: What is the relationship between the 'User', 'Website', and 'Server'? **Answer**: The 'User' accesses the 'Website' through a 'Blog'. The 'Website' points to the 'Server', which is hosting the 'Blog'. |
| Fig. 7. A diagram explaining virtual-physical address translation in computer memory. | **Question**: What is the relationship between the virtual page number, physical page number, and page offset? **Answer**: The virtual page number, physical page number, and page offset are all related by numbers that are stored in the page table. The page table entry tells us which page has which virtual page number and physical page number. |
| Fig. 8. A diagram showing the structure of the cell. | **Question**: What is the function of the Golgi complex? **Answer**: The Golgi complex is responsible for organizing and shipping proteins within the cell. |

### C. Reinforcement Learning

Our RL pipeline, illustrated in Figure 4, includes following components:

- Policy Model: the model being optimized (in our case, T5).
- Reference model: a fixed copy of the actor prior to PPO updates, used to compute per-token KL penalties.
- Reward model: a cross-encoder trained on preference data, providing scaler signals for generated questions.

We have chosen DistillRoBERTa as the reward model for our reinforcement learning pipeline, which is a variant of RoBERTa [14] that has been distilled using the same approach used to create DistillBERT [15]. We initially intended to fine-tune it using human feedback on questions generated by our system. However, collecting and reviewing the amount of data required to make a tangible difference in the model's performance was infeasible given our project constraints. To circumvent this, we attempted to use AI-generated feedback for training the reward model, but the results were not satisfactory. We have gone forward with using a variant of DistillRoBERTa that has been pre-trained for natural language understanding (NLU) generally – that is, agnostic to any particular task or dataset. We experimented with the model as-is, without any fine-tuning, and noted a slight increase in question scores (BLEU and ROUGE) and a slight drop in answer scores, as shown in Table XI.

TABLE XI. RESULTS BEFORE AND AFTER REINFORCEMENT LEARNING

| Metric | Questions Before RL | Questions After RL | Answers Before RL | Answers After RL |
|---|---|---|---|---|
| BLEU-4 | 0.1230 | 0.1237 | 0.3150 | 0.3141 |
| ROUGE-L | 0.3957 | 0.3967 | 0.7426 | 0.7404 |

Table XII shows sample outputs on the same input context, both before and after reinforcement, demonstrating some improvement in question structure and quality.

TABLE XII. OUTPUT BEFORE AND AFTER REINFORCEMENT LEARNING

| Context | Before RL | After RL |
|---|---|---|
| Photosynthesis is the process by which green plants, algae, and some bacteria convert light energy into chemical energy. This process occurs mainly in the chloroplasts of plant cells and involves several key steps: 1. Light Absorption: Chlorophyll in the plant cells absorbs sunlight. 2. Water Splitting: Water molecules are split into oxygen and hydrogen. 3. Carbon Dioxide Fixation: Carbon dioxide from the air is used to produce glucose. 4. Oxygen Release: Oxygen is released as a byproduct | **Question**: What property of oxygen does chlorophyll in plant cells absorb? **Answer**: 4. light | **Question**: In what type of plant cells does photosynthesis occur? **Answer**: chloroplasts |
| Reinforcement Learning from Human Feedback (RLHF) is a technique used to align language models with human preferences. It typically involves three steps: supervised fine-tuning of a pre-trained model, training a reward model based on human-labeled comparisons, and finally, optimizing the language model using reinforcement learning algorithms such as Proximal Policy Optimization (PPO). [...] | **Question**: The final step involves what? **Answer**: optimizing the language model using reinforcement learning algorithms such as Proximal Policy Optimization (PPO) | **Question**: What property of oxygen does chlorophyll in plant cells absorb? **Answer**: fine-tuning of a pre-trained model, training a reward model based on human-labeled comparisons, and finally, optimizing the language model |

We speculate that the model's generalized pretraining alone is somewhat sufficient for evaluating our model and generating feedback scores. However, fine-tuning it on user feedback would certainly result in significant improvement in question alignment with human preferences.

## V. CONCLUSION

Our work constitutes an end-to-end system for multimodal question generation, balancing functionality, resource efficiency and user experience. We have demonstrated the capabilities of both T5 as a question generation model across different question types, and ChartGemma as a visual question generator (a task it had not been tested for before). We have also set up a production-ready pipeline for reinforcement learning from human feedback to ensure that the question generation process improves iteratively as users rate the output.

Due to the breadth of the project, there is a large surface for optimization and innovation. The most notable area for improvement is the collection of question feedback, through the system or otherwise, and using this data to reinforce the question generation model. This will serve as a crowdsourced dataset to improve the system's capabilities and also serve as a resource for further work in the field.

As for additional features, an interesting one would be allowing the user to upload sample questions or past exam papers, and then generating new questions that follow similar patterns. Also, introducing adaptive difficulty levels will make the model more valuable as an assessment tool. Finally, support for multilingual question generation will allow a wider demographic to benefit from the system. The system can also be extended into a full-featured platform with student dashboards and performance analytics. Quiz data can be used to customize a study plan, and identify weakness points. Questions can then be generated to specifically target these weakness points, helping the students polish their skills and build confidence in trouble areas.

This project lays the groundwork for multimodal question generation, and is our contribution to the growing research at the intersection of AI and education, a field with brilliant potential and far-reaching benefits.

ACKNOWLEDGMENT

We extend our utmost gratitude to Allah, the most Gracious and Compassionate. It is through His grace that we have been able to embark on this journey and persevere through its challenges.

We sincerely thank Dr. Ahmed Salah, our supervisor, whose guidance was invaluable throughout this project's journey – from its earliest moments as a hopeful idea to its full realization. He has been an unwavering support for our team, both morally and technically, and we are honoured to have worked with him.

We thank TA Youssef Sobhy for supervising and supporting our project. We also thank everyone who has contributed to bringing this project to life, from faculty staff to fellow students, whether by suggesting ideas, sharing resources and experiences, or cheering for us during difficult times. You are a wonderful community that we are grateful to be a part of.

Finally, we wholeheartedly thank our families, who have supported us through thick and thin – who witnessed our sleepless nights, joyful celebrations and everything in between. This wouldn't have been possible without you, and this achievement is yours to celebrate just as much as it is ours.